\documentclass[sigconf]{acmart}
\AtBeginDocument{%
  }

\setcopyright{acmlicensed}
\copyrightyear{2018}
\acmYear{2018}
\acmDOI{XXXXXXX.XXXXXXX}
\acmConference[Conference acronym 'XX]{Make sure to enter the correct
  conference title from your rights confirmation email}{June 03--05,
  2018}{Woodstock, NY}
\acmISBN{978-1-4503-XXXX-X/2018/06}

\usepackage{color}           % 定义基本颜色
\usepackage{xcolor}          % 扩展颜色定义
\usepackage{textcomp}        % 支持文本符号（如上下标）
\usepackage{tcolorbox}
\usepackage{pifont}

\usepackage{array}           % 表格列类型支持
\usepackage{booktabs} % Professional-looking tables (optional but recommended)
\usepackage{multirow} % Enables row merging
\usepackage{multicol}        % 支持单元格跨列
\usepackage{adjustbox}       % 表格缩放和居中
\usepackage{dsfont}

\usepackage{xspace}

\newcommand{\datasetFont}[1]{\textnormal{#1}}

\newcommand{\oursdata}{\datasetFont{ShadeBench}\xspace}

\begin{document}

\title{ShadeBench: A Benchmark Dataset for Building Shade Simulation in Sustainable Society}

\author{Longchao Da$^1$, Mithun Shivakoti$^1$, Xiangrui Liu$^1$, T Pranav Kutralingam$^1$, Yezhou Yang$^1$, Hua Wei$^{1,2}$}

\affiliation{%
  \institution{$^1$ School of Computing and Augmented Intelligence, Arizona State University}
  \institution{$^2$ Global Futures Laboratory, Arizona State University}
  \city{Tempe}
  \state{AZ}
  \country{USA}
}

\renewcommand{\shortauthors}{Longchao Da, et al.}

\begin{abstract}
Urban heat exposure is becoming an increasingly critical challenge due to the intensifying urban heat island effect. Fine-grained shade patterns, especially those induced by urban buildings, strongly influence pedestrians’ thermal exposure and outdoor activity planning. However, accurately modeling and analyzing urban shade at scale remains difficult because of the lack of large-scale datasets and systematic evaluation frameworks.
To address this challenge, we present ShadeBench, a comprehensive dataset and benchmark for urban shade understanding. ShadeBench contains geographically diverse urban scenes with temporally varying simulated shade maps and textual descriptions, together with aligned satellite imagery, building skeleton representations, and 3D building meshes.
Built upon this multimodal dataset, ShadeBench supports a range of downstream tasks, including shade generation, shade segmentation, and 3D building reconstruction. We further establish standardized evaluation protocols and baseline methods for these tasks.
By enabling scalable and fine-grained shade analysis, ShadeBench provides a foundation for data-driven urban climate research and supports future studies in heat-resilient urban planning and decision-making. The code and dataset are publicly available at \text{\url{https://darl-genai.github.io/shadebench/}}.
\end{abstract}


% \ccsdesc[500]{Computing methodologies~Image-based rendering}
% \ccsdesc[100]{Computing methodologies~Supervised learning by regression}


\received{8 February 2026}

\maketitle

\section{Introduction}
% Heatwaves are becoming increasingly frequent and severe worldwide~\cite{meehl2004more}, which poses a growing threat to public health and urban sustainability. Recent reports indicate that both the intensity and duration of extreme heat events have risen markedly over the past decades, leading to substantial increases in heat-related mortality~\cite{casereport,monashstudy}.\begin{figure}[h!]
%     \centering
%     \includegraphics[width=0.99\linewidth]{figs/hookFigure.pdf}
%     \caption{Growing risks of extreme urban heat revealed by a case in Arizona where the temperature exceeds 50 °C (122 °F). The \oursdata leverages a comprehensive dataset and systematic evaluation protocols to support shade-aware urban analysis under realistic conditions, including shade simulation, shade segmentation, and 3D building reconstruction.}
%     \label{fig:intro}
% \end{figure} The World Health Organization has identified extreme heat as one of the leading causes of weather-related deaths globally, with great impacts on vulnerable populations (such as older adults and individuals who live or work in dense urban environments~\cite{niehsheatimpact}). In regions such as the southwest United States, the severity of this issue is particularly pronounced: during summer, temperatures in Arizona have reached 128°F (53.3°C), with prolonged periods exceeding 122°F (50°C). As shown in Figure~\ref{fig:intro}, these extreme conditions are not merely numerical records, they directly endanger human health, disrupt outdoor activities, and threaten the safety of the essential outdoor laborers.

\begin{figure}[h!]
    \centering
    \vspace{-3mm}
    \includegraphics[width=0.99\linewidth]{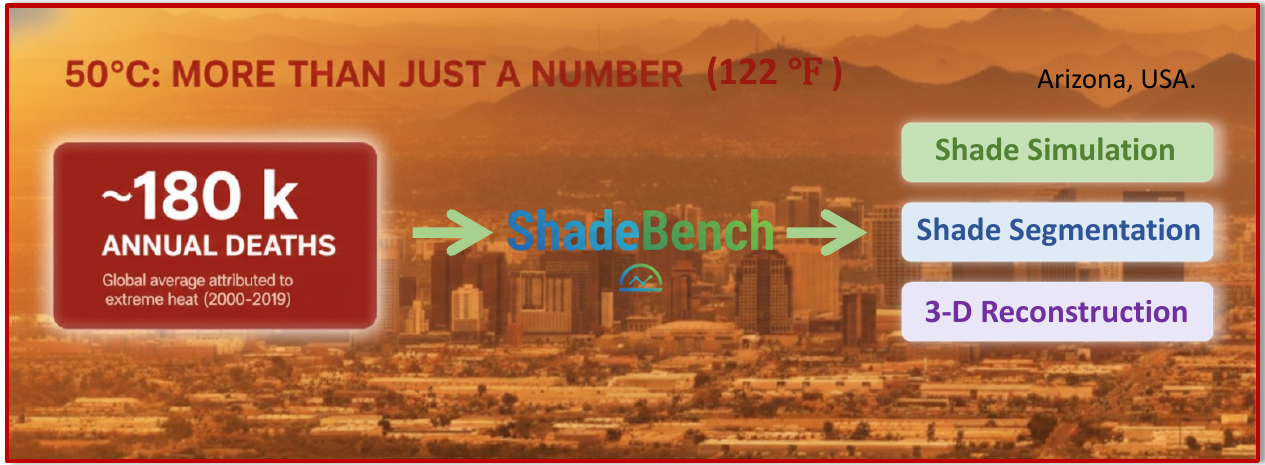}
    \vspace{-3mm}
    \caption{Growing risks of extreme urban heat revealed by a case in Arizona where the temperature exceeds 50 °C (122 °F). \oursdata leverages a comprehensive dataset and systematic evaluation protocols to support shade-aware urban analysis under realistic conditions, including shade simulation, shade segmentation, and 3D building reconstruction.}
    \label{fig:intro}
    \vspace{-3mm}
\end{figure}

Heatwaves are becoming increasingly frequent and severe worldwide~\cite{meehl2004more}, which poses escalating challenges to public health systems and the long-term sustainability of urban environments. Recent large-scale assessments indicate that both the intensity and duration of extreme heat events have increased substantially over recent decades, leading to a significant rise in heat-related mortality~\cite{casereport,monashstudy}.
The World Health Organization has identified extreme heat as one of the leading causes of weather-related mortality worldwide, disproportionately affecting vulnerable populations such as older adults and individuals who live or work in dense urban environments~\cite{niehsheatimpact}. This risk is particularly pronounced in the southwestern United States, where summer temperatures in Arizona have reached 128°F (53.3°C), with extended periods exceeding 122°F (50°C). As illustrated in Figure~\ref{fig:intro}, such extreme thermal conditions are not merely statistical extremes but represent tangible and recurring hazards that directly threaten human health, constrain outdoor activity, and endanger essential outdoor labor forces.

Shade plays a critical role in mitigating heat exposure by reducing direct solar radiation and surface heating~\cite{guo2023canopy}. In urban environments, the availability and spatiotemporal distribution of shade are largely determined by built infrastructure, particularly building geometry and layout, which govern patterns of solar occlusion throughout the day and across seasons~\cite{ahmadian2021effect}. Prior studies have shown that explicitly incorporating shade information into urban planning and routing systems can reduce individual heat exposure and associated health risks~\cite{ma2018parasol,da2024shaded}. Despite its demonstrated importance, existing approaches to urban shade analysis exhibit several fundamental limitations: \ding{182}~Many synthetic datasets rely on simulation-based ground-truth shade~\cite{da2025deepshade} derived from building information maintained in open-source resources such as OpenStreetMap (OSM). Because OSM data are often outdated or incomplete, these datasets frequently exhibit misalignment between simulated building geometry and real-world urban layouts observed in satellite imagery, which can affect model training and evaluation. Beyond geometric misalignment, prior datasets~\cite{da2025deepshade} neglect geographic variability in solar trajectories, implicitly assuming similar sun-path behavior across locations. In particular, well-established differences in solar motion between the Northern and Southern Hemispheres induce distinct diurnal and seasonal shade evolution patterns, which are not precisely modeled, thereby limiting the physical realism of the simulated shade data. \ding{183}~existing simulation-based baselines are rarely assessed under unified and standardized evaluation protocols~\cite{yang2023diffusion, da2025deepshade, zhang2023adding}, making it difficult to rigorously compare accuracy, robustness, and the ability to capture realistic shade dynamics. \ding{184}~the absence of standard benchmarks constrains support for downstream shade-related tasks, including generative shade simulation, satellite-based shade segmentation, and 3D building reconstruction, all of which are essential for actionable urban analysis and heat-aware planning~\cite{da2024shaded}.

To address these challenges, we introduce \textbf{\oursdata}, a comprehensive benchmark for large-scale urban shade understanding. For challenge \ding{182}, we develop a rigorous construction pipeline to address mismatches between simulation-derived shade ground truth and real-world observations. We identify discrepancies caused by missing or outdated building data and viewpoint-dependent observation biases. In addition, to overcome the simplified or location-agnostic assumptions of solar movement, we therefore develop a missing-building handling strategy that preserves valid structures while detecting and masking non-existent or misaligned regions at the pixel level, and apply the NOAA Solar Position Algorithm~\cite{reda2004solar} to ensure physically consistent sun-driven shade simulation. 
% This process yields shade annotations aligned with real-world urban topology across diverse observation conditions. 
For challenge \ding{183},
Based on the constructed dataset, we establish a unified and systematic evaluation protocol to benchmark existing simulation-based shade modeling approaches. Specifically, we evaluate Diffusion model~\cite{yang2023diffusion}, ControlNet~\cite{zhang2023adding}, and DeepShade~\cite{da2025deepshade} methods under consistent geographic, temporal, and urban-layout settings, enabling quantitative assessment of their accuracy, robustness, and comparative advantages in capturing realistic shade dynamics.
For challenge \ding{184}, we leverage this dataset and establish standardized benchmarks that support three downstream tasks: including shade generation evaluation, benchmarking segmentation models for accurate delineation of complex urban shadow structures, and assessing representative 3D building reconstruction methods to examine how reconstruction quality propagates to downstream shade estimation performance.% Together, this benchmark enables systematic comparison, reproducibility, and deeper analysis of model behavior across shade-related tasks.
In summary, the contributions of this paper are as follows: 

\noindent$\bullet$~We introduce a large-scale urban shade dataset with a rigorous construction pipeline that aligns simulation shade ground truth with real observations, meanwhile incorporating location-dependent solar trajectories to ensure physically realistic urban shade dynamics. 

\noindent$\bullet$~We establish a unified and standardized evaluation protocol for benchmarking simulation-based shade modeling methods, enabling systematic comparison of accuracy, robustness, and ability to capture temporal dynamics.

\noindent$\bullet$~We provide benchmark support for multiple downstream shade-related tasks, including shade generation, shade segmentation, and 3D building reconstruction.
% \vspace{-1mm}
% \begin{itemize}
%     \item \textbf{First}, we introduce a large-scale urban shade dataset with a rigorous construction pipeline that aligns simulation shade ground truth with real observations, meanwhile handling missing and outdated building geometry and incorporating location-dependent solar trajectories to ensure physically realistic urban shade dynamics.
%     \item \textbf{Second}, we establish a unified and standardized evaluation protocol for benchmarking simulation-based shade modeling methods, enabling systematic comparison of accuracy, robustness, and ability to capture temporal dynamics.
%     \item \textbf{Third}, we provide benchmark support for multiple downstream shade-related tasks, including shade generation, shade segmentation, and 3D building reconstruction.
% \end{itemize}
Together, our work positions as a foundation for shade-aware urban analytics and supports future research on heat-resilient urban planning and decision-making.

\section{Related Work}
To the best of our knowledge, existing studies lack a comprehensive benchmark that jointly supports aligned shade data, standardized evaluation of simulation baselines, and systematic assessment of downstream shade-related tasks. As a result, we organize the related work into three complementary perspectives:

\begin{figure*}[t!]
    \centering
    \includegraphics[width=0.95\textwidth]{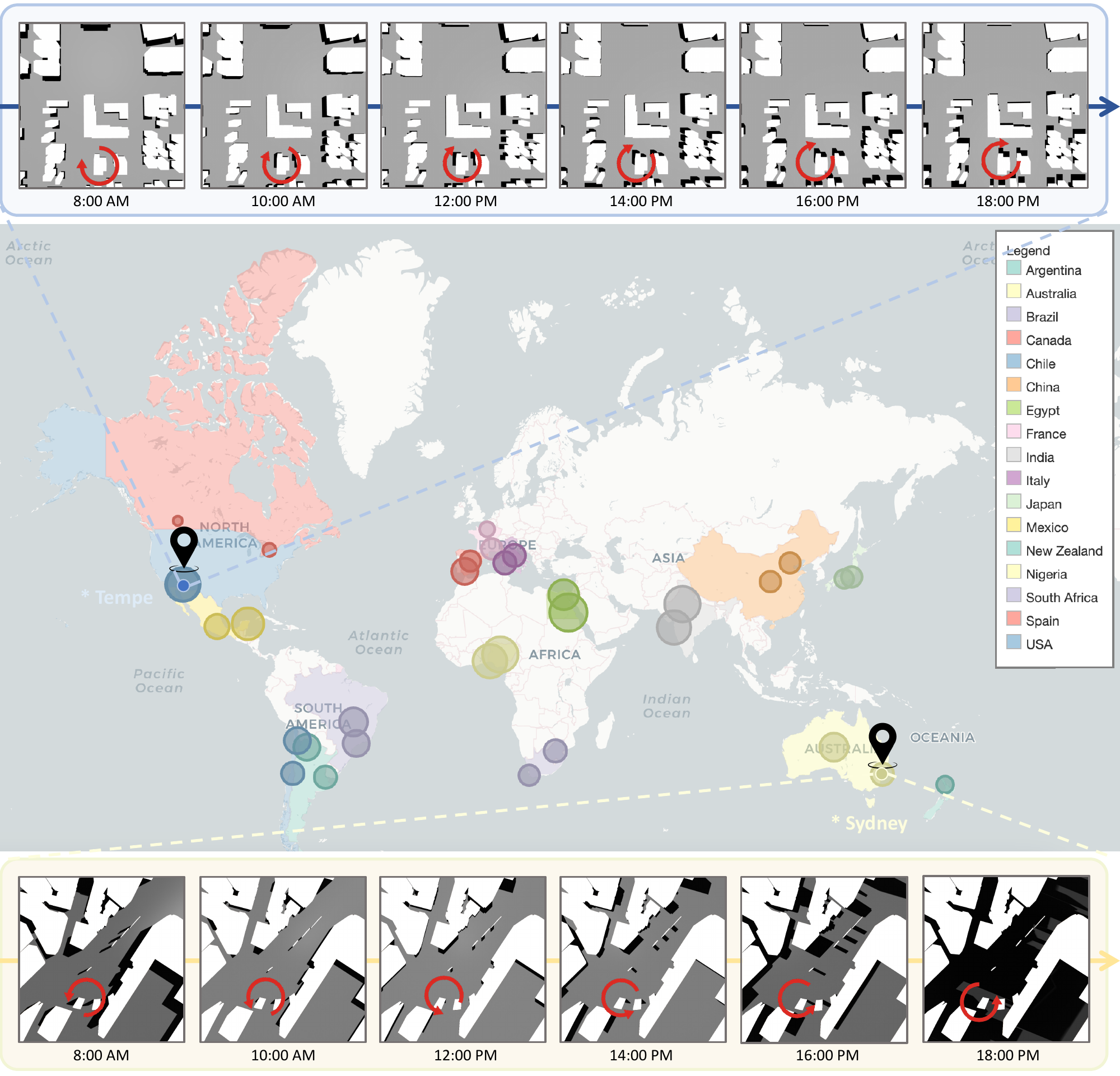}
    \vspace{-3mm}
    \caption{Global coverage and hemispheric consistency of the \oursdata dataset. \oursdata covers 6 continents and 16 countries, with 34 cities, each country featuring two representative urban layouts (dense and sparse). Circle size indicates relative sun-exposure severity. The top and bottom panels illustrate diurnal shade evolution in two example cities from opposite hemispheres: Tempe, Arizona (USA; Northern Hemisphere, top) and Sydney, Australia (Southern Hemisphere, bottom). Following the red-circled direction, the temporal progression reveals opposite shade-rotation patterns between the two hemispheres, reflecting the reversal of solar trajectories across the equator. This hemispheric contrast confirms that the dataset adheres to geographic rules, capturing physically consistent, location-aware shade dynamics across the globe. }
    \label{fig:worldmap}
\end{figure*}

\vspace{1mm}
\noindent \textbf{Building Shade Analysis.}
A growing body of research has investigated the role of building-induced shade in shaping urban thermal conditions and human well-being. For instance, the work~\cite{park2023quantifying} analyzed the aggregated cooling benefits of urban shade, demonstrating the value of high-resolution thermal observations in hot and arid contexts such as Tempe, Arizona. Other studies have employed physically based models, including SOLWEIG~\cite{lindberg2008solweig}, to characterize the localized thermal effects of shade with fine spatial detail, offering actionable insights for urban design. These research efforts have further motivated practical interventions such as “cool corridor” initiatives~\cite{buo2023high, wu2020construction}, which seek to reduce pedestrian heat exposure through strategic enhancement of sidewalk shade. While these approaches have advanced urban heat mitigation strategies, they are largely conducted in localized settings and tailored to specific cities or neighborhoods. This localization makes it difficult to derive globally applicable insights or to systematically compare shade-related findings across diverse urban environments. More importantly, the lack of a comprehensive and standardized dataset for urban shade analysis limits scalability and cross-city evaluation. This motivates the need for a large-scale benchmark to support consistent and generalizable shade analysis.

\vspace{1mm}
\noindent \textbf{City Simulation.}
High-resolution urban datasets have also been employed to support detailed shade analysis. A representative example is the University of Texas GLObal Building Heights for Urban Studies (UT-GLOBUS) dataset~\cite{li2020developing}, which fuses information from multiple sensing sources, including LiDAR, ICESat-2~\cite{abdalati2010icesat}, and the Global Ecosystem Dynamics Investigation (GEDI)~\cite{schneider2020towards}, together with machine learning techniques to produce large-scale urban canopy representations~\cite{kamath2024global}. In parallel, simulation-based studies using geographic information systems (GIS), such as the analysis conducted in the Adab neighborhood of Sanandaj~\cite{beheshtifar2023simulation}, demonstrate the challenges of accurately modeling shade in dense, high-rise urban settings by leveraging 3D GIS data to simulate seasonal variations in shading patterns. Despite these advancements, existing datasets and simulation frameworks still suffer from misalignment between modeled urban geometry and real-world observations, limiting their reliability for downstream analysis. Besides, the techniques supported by these datasets are often narrowly scoped, typically tailored to either simulation or case analysis, but not jointly enabling simulation, planning, large-scale urban analysis, and model learning within a unified framework. This fragmentation highlights the need for a comprehensive and aligned benchmark dataset that can support diverse shade-related tasks across varied urban environments.

\begin{figure*}[t!]
    \centering
    \includegraphics[width=1\textwidth]{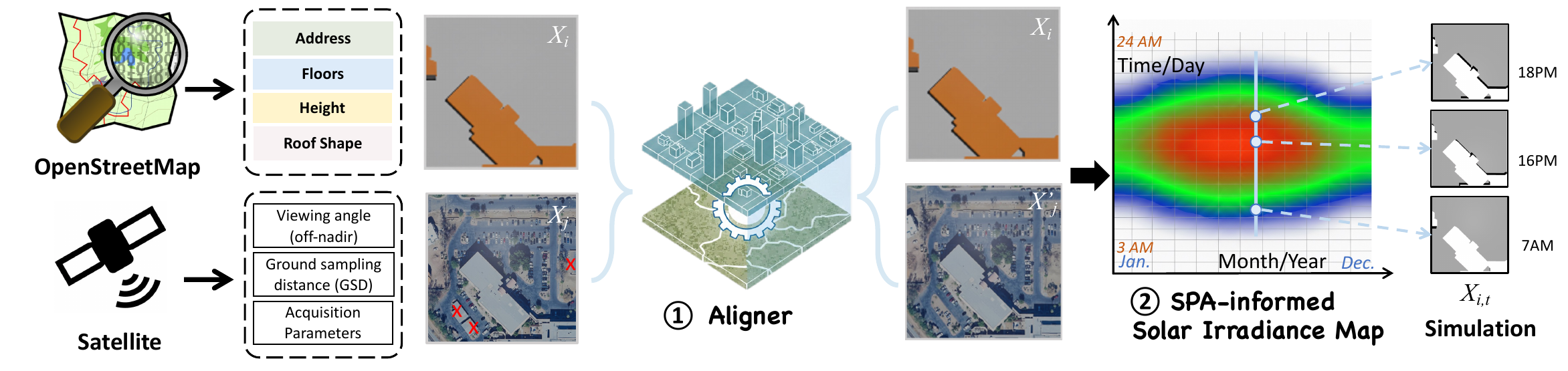}
    \vspace{-6mm}
   \caption{Overview of the dataset construction pipeline with alignment and physically grounded solar modeling. The pipeline integrates map-based building priors from OpenStreetMap and sensor-based satellite observations, which naturally exhibit geometric misalignment due to missing or outdated building information and acquisition-related biases. We denote the simulated building layout derived from OpenStreetMap as $X_i$ and the corresponding satellite observation as $X_j$. \ding{172} \textbf{Aligner} reconciles these inputs by detecting, correcting, and masking non-existent or misaligned structures, producing spatially consistent representations $(X_i', X_j')$. \ding{173} \textbf{SPA-informed Solar Irradiance Map} computes location- and time-dependent solar irradiance using physically grounded solar trajectories, enabling realistic shade simulation across different hours and seasons.
The resulting shade observation at time $t$ is denoted as $X_{i,t}$.
}
    \label{fig:pipeline}
\end{figure*}
% \vspace{-1mm}

\vspace{1mm}
\noindent \textbf{Generative Models.}
Recent advances in generative modeling have shown strong performance across a range of visual synthesis tasks. In particular, Generative Adversarial Networks~\cite{brock2018large} and diffusion-based models~\cite{croitoru2023diffusion} have achieved notable success in image generation~\cite{dhariwal2021diffusion} and image refinement~\cite{du2023arsdm}. Complementary architectures such as ControlNet~\cite{zhao2024uni} and StyleGAN~\cite{abdal2019image2stylegan} further enhance controllability and realism in synthetic image generation, where ControlNet incorporates external structural cues for spatial conditioning and StyleGAN supports high-resolution synthesis with fine-grained style manipulation~\cite{azadi2018multi}. Despite these advances, only a small number of studies have explored the application of generative models to urban shade analysis~\cite{da2025deepshade}. However, existing datasets used in these works often suffer from misalignment between simulation-derived building geometry and real-world satellite imagery, which can affect model learning and evaluation. Besides, there is a lack of standardized evaluation protocols for shade-related tasks such as shade simulation, segmentation, and 3D building reconstruction. This hinders the objective assessment of simulation methods and restricts the applicability across diverse urban environments. In contrast, our work introduces a unified benchmark that provides aligned data and standardized evaluation across multiple shade-related tasks, enabling rigorous and reproducible analysis of generative models for urban shade understanding.

\section{The \textbf{\oursdata} Dataset}
\begin{figure}[h!]
    \centering
    \includegraphics[width=0.45\textwidth]{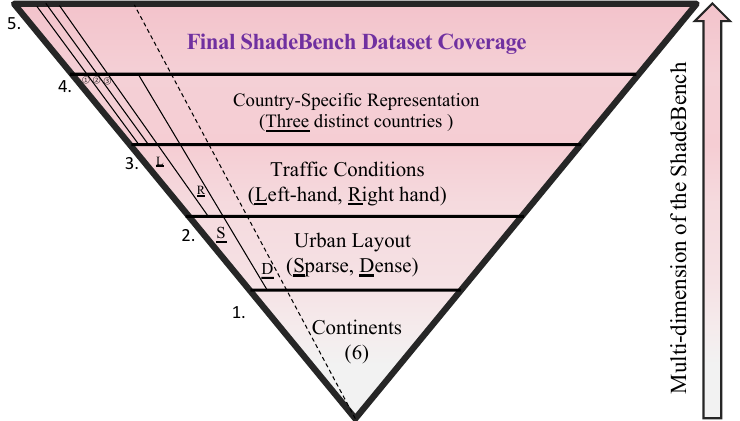}
    \vspace{-2mm}
    \caption{The design rationale of \oursdata.}
    \label{fig:dims}
    \vspace{-5mm}
\end{figure}

In this section, we present a pipeline for constructing a comprehensive dataset for reliable urban shade understanding that provides rich modalities, aligned simulation and real-world building footprints, and comprehensive data resources. 
As shown in Figure~\ref{fig:dims}, \oursdata spans six continents, covering diverse urban layouts (sparse vs. dense), and traffic rules (left-hand vs. right-hand driving), and includes at least three distinct countries per category to ensure broad geographic and infrastructural representation.
In the following subsections, we introduce the detailed pipeline, as shown in Figure~\ref{fig:pipeline}.

\subsection{Simulation Setup}
Leveraging the OpenStreetMap (OSM), we retrieve metadata required for large-scale urban shade simulation. Given the geographic coordinates (latitude and longitude) of each area of interest, we extract building attributes including address, height, number of floors, and roof shape as shown in the top-left of Figure~\ref{fig:pipeline}.
These attributes collectively define the 3D building topology, represented as a footprint-based skeleton suitable for simulation. Based on the retrieved metadata, we perform shade simulation using Blender~\cite{gschwandtner2011blensor}. We construct a city-scale simulation environment by extruding building footprints according to their geometric and height attributes. For each area of interest, the simulated scene is rendered at a fixed spatial resolution aligned with standard web map tiling, specifically corresponding to Google Maps tile level~13.
Details of the resolution mapping and scale calibration are in Appendix~\ref{sec:reso}.

\subsection{Alignment}

\subsubsection{Geometric Aligner} 
A key challenge in dataset construction is the geometric misalignment between simulated urban layouts and real-world satellite observations. Such misalignment commonly arises when OSM metadata is outdated, resulting in buildings that appear in satellite imagery but are absent from the simulated environment. To address this challenge,
we design an aligner to detect and correct regions in a satellite image that are inconsistent with the simulated urban geometry.
Let $X_i$ denote the simulated building topology (skeleton image) built from OpenStreetMap metadata, and let $X_j$ denote the corresponding satellite image, the detection and correction steps are as follows:

\vspace{1mm}
\noindent $\bullet$~\textbf{Edge-based inconsistency detection.}
We first extract structural edges from the satellite image $X_j$ using a Canny edge detector~\cite{maranga2025learned}, yielding a binary edge map $E_j = \mathcal{\text{Can}}(X_j)$, where $\mathcal{\text{Can}}(\cdot)$ is the Canny edge extraction operator.
Similarly, we derive the simulated building contour map from the skeleton image $X_i$ as $E_i = \mathcal{B}(X_i)$,
where $\mathcal{B}(\cdot)$ extracts building boundaries from the simulated topology. Then, we define a \emph{building context inconsistency map} $C_j$ by identifying edge responses in the satellite image that are unsupported by the simulated geometry:
$
    C_j = E_j \cdot \mathbb{I}(E_i = 0)
$
where $\mathbb{I}(\cdot)$ is the indicator function.
Pixels with non-zero values in $C_j$ correspond to building-like structures present in the satellite observation but absent from the simulation, indicating potential misalignment due to outdated or missing building metadata.

\vspace{1mm}
\noindent $\bullet$~\textbf{Alignment via structural correction.}
Given the detected inconsistency map $C_j$, we propose to reconcile the misalignment using the following two strategies. Both strategies aim to produce a refined satellite image $X_j'$ that is geometrically consistent with the simulated topology $X_i$:
\ding{182}~Mask-based alignment:
In this approach, we directly suppress inconsistent regions by masking them out using the simulated building support:
$
    X_j' = X_j \cdot \mathbb{I}(X_i > 0)
$
This operation removes satellite-observed structures that do not correspond to any building in the simulation, enforcing strict geometric consistency.
\ding{183}~Generative refinement:
Alternatively, we apply a generative image editing function $\mathcal{G}(\cdot)$ to inpaint inconsistent regions into semantically plausible non-building surfaces:
\begin{equation}
   X_j' = \mathcal{G}(X_j, C_j) \label{eq:genImage} 
\end{equation}   
where $\mathcal{G}$ is instantiated using a GenAI-based image editing model (e.g., Nano-banana~\cite{qian2025pico}).
In this formulation, regions indicated by $C_j$ are replaced with contextually consistent alternatives such as parking lots or open ground, preserving visual continuity while removing unsupported building structures.
After alignment, we obtain a pair of spatially consistent representations $(X_i, X_j')$, where $X_j'$ is the refined satellite image. An example is provided in the Figure~\ref{fig:pipeline}. These aligned outputs serve as the geometric foundation for subsequent solar and shade simulations.

\subsubsection{SPA-informed Solar Irradiance Map}
Beyond geometric alignment, realistic shade simulation requires physically consistent modeling of solar movement.
Prior datasets~\cite{da2025deepshade} often rely on simplified or location-agnostic assumptions of sun trajectories, which fail to capture hemispheric differences in solar behavior.
To capture hemispheric differences, we explicitly model the sun's apparent position using the NOAA Solar Position Algorithm (SPA)~\cite{reda2004solar}, which provides physically grounded solar geometry for arbitrary locations and times.
Given geographic latitude $\phi$, longitude $\lambda$, calendar date $d$, and time of day $t$, the SPA computes the solar elevation angle $\alpha(d,t)$ and solar azimuth angle $\psi(d,t)$. These quantities jointly define the unit sun direction vector as: 
\begin{equation}
    \mathbf{s}(d,t) =
    \begin{bmatrix}
        \cos\alpha(d,t)\cos\psi(d,t) \\
        \cos\alpha(d,t)\sin\psi(d,t) \\
        \sin\alpha(d,t)
    \end{bmatrix}\label{eq:solar}
\end{equation}
which specifies the incident direction of solar rays in the local geographic coordinate system.
This vector is used to parameterize the directional light source in the Blender-based simulation environment. Based on the sun direction $\mathbf{s}(d,t)$, we define an SPA-informed solar irradiance map $I(d,t)$ that characterizes the relative solar exposure as a function of time and season:
$
    I(d,t) = g\!\left(\alpha(d,t)\right)
$
where $g(\cdot)$ denotes a physically motivated irradiance function that monotonically increases with solar elevation and encodes diurnal and seasonal variation in solar intensity.

During rendering, the simulated urban geometry associated with building topology $X_i$ is illuminated by the directional light $\mathbf{s}(d,t)$, producing a time-dependent shade observation as: 
\begin{equation}
        X_{i,t} = \mathcal{R}\!\left(X_i, \mathbf{s}(d,t)\right)\label{eq:R}
\end{equation}
where $\mathcal{R}(\cdot)$ denotes the rendering operator that accounts for geometric occlusion and surface projection. By varying $(d,t)$, this formulation enables the generation of physically consistent shade images across different hours of the day and months of the year, capturing realistic temporal and seasonal shade dynamics.

In addition, hemispheric differences in shade evolution are explicitly captured through the geographic latitude parameter $\phi$ in the solar position formulation. In the NOAA Solar Position Algorithm, both the solar elevation $\alpha(d,t)$ and azimuth $\psi(d,t)$ are functions of $(\phi, \lambda, d, t)$. When the sign of $\phi$ changes across the equator, the resulting azimuth trajectory $\psi(d,t)$ follows an opposite diurnal progression, which in turn produces a mirrored sun direction vector $\mathbf{s}(d,t)$ as defined in Eq.~(\ref{eq:solar}). This sign-dependent variation propagates through the rendering operator $\mathcal{R}(\cdot)$ in Eq.~(\ref{eq:R}), leading to opposite shadow rotation patterns for identical urban geometries in the Northern and Southern Hemispheres, as illustrated in Figure~\ref{fig:worldmap}.

\subsection{Data Collection}\label{sec:data}

\begin{figure}[t!]
    \centering
    \includegraphics[width=0.48\textwidth]{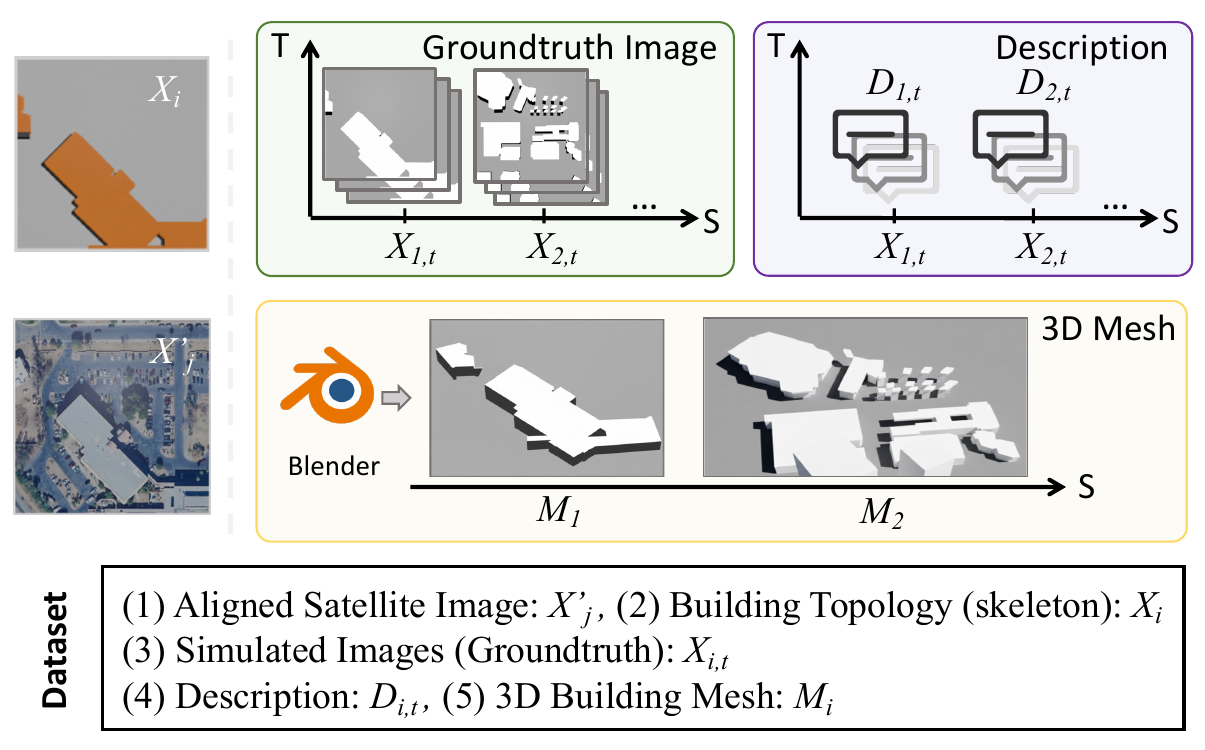}
    \vspace{-6mm}
   \caption{ Data components contained in \oursdata. Each dataset sample is constructed from the aligned outputs of the pipeline in Figure~\ref{fig:pipeline} and consists of multiple complementary modalities. Given an aligned satellite image $X_j'$ and its corresponding building topology (skeleton) $X_i$, we provide a sequence of simulated shade images $\{X_{i,t}\}$ as ground-truth annotations, where $t\in T$ indicates temporal variations, and $S$ captures both spatial differences. Every $X_{i,t}$ is paired with a structured textual description $D_{i,t}$ with solar attributes. For each spatial tile, we also include a Blender-exported 3D building mesh $M_i$ to represent the urban geometry and enable downstream geometry-aware analysis and evaluation.
}
    \label{fig:consist}
    \vspace{-3mm}
\end{figure}

After simulation, alignment, and solar modeling are completed, we collect the final dataset components as illustrated in Figure~\ref{fig:consist}.
For each spatial tile, the dataset includes:
(1) an aligned satellite image $X_j'$,
(2) a building topology (skeleton) image $X_i$,
(3) a sequence of simulated shade ground-truth images $X_{i,t}$ capturing spatial and temporal variation,
(4) structured textual descriptions $D_{i,t}$ encoding solar attributes such as declination, elevation, and time, and
(5) a Blender-exported 3D building mesh $M_i$ in  \texttt{Wavefront} \texttt{.OBJ} format.

\begin{figure*}[t!]
    \centering
    \includegraphics[width=1\linewidth]{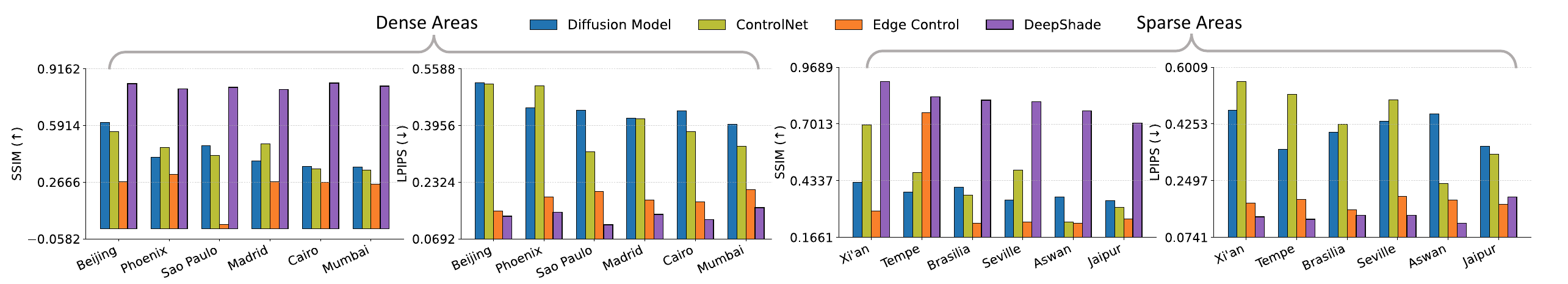}
    \vspace{-4mm}
    \caption{Performance of generative simulation under two scenario types (dense\textbf{*} and sparse\textbf{*} building cities) w.r.t. SSIM ($\uparrow$) and LPIPS ($\downarrow$). Across both scenarios, methods that explicitly incorporate geometric structure and conditioning signals produce more stable and perceptually consistent shade patterns (DeepShade). Performance gaps become more obvious in sparse areas, where long shadows and weak occlusions increase sensitivity to solar conditions and urban layout.}
    \label{fig:result1}
    % \vspace{-2mm}
\end{figure*}

These components form a unified multimodal dataset that supports shade simulation, segmentation, and geometry-aware tasks such as 3D building reconstruction under realistic urban and environmental conditions. As part of data item (4), we provide structured textual descriptions to enable condition-based and controllable shade generation. Each text description is formulated as
$
    T = f(\theta_{sun}, t_{day})
$
where \( T \) denotes the generated text prompt, \( \theta_{sun} \) represents the solar declination or elevation angle, and \( t_{day} \) corresponds to the time of day. The function \( f(\cdot) \) maps physical solar and temporal attributes into natural language expressions through a lightweight string composition scheme with multiple descriptive variants. These text prompts can be used independently or jointly with visual inputs to condition downstream generative or predictive models.
The following text prompts illustrate representative text descriptions under different temporal conditions:
\vspace{-2mm}
\begin{tcolorbox}[colback=gray!10,
                  colframe=black,
                  width=8.5cm,
                  arc=1mm, auto outer arc,
                  boxrule=0.05pt,
                  fontupper=\small]
\vspace{-1mm}
% \textbf{Example Text Prompts:} \\[2pt]
\textbf{Prompt 1:} Solar declination: $-20.7$ \\ 
\textbf{Prompt 2:} Solar elevation angle: $45^\circ$ \\ 
\textbf{Prompt 3:} Local time: 6:00 PM
\vspace{-1mm}
\end{tcolorbox}
\vspace{-2mm}

Overall, the dataset construction ensures spatial alignment between simulated and real-world observations while supporting controllable shade generation via image inputs, textual conditioning, or flexible combination. Although the urban environments are modeled using \texttt{Blender} and OSM data, missing and outdated structures are explicitly handled to facilitate robust model training and generalization to unseen domains. The dataset is split into training, validation, and test sets with a 60\%/20\%/20\% ratio to enable standardized benchmarking. Additional details are in Appendix~\ref{sec:datasetdetail}.

\section{The \oursdata Tasks}
% \textcolor{red}{Benchmarking tasks description}
Based on the aligned multimodal dataset and physically grounded shade simulation pipeline, \oursdata supports a set of standardized benchmarking tasks that evaluate urban shade understanding from complementary perspectives, including generative shade simulation, shade segmentation, and 3D reconstruction.

\subsection{Generative Shade Simulation}

% \textbf{Task Objective.}
The generative shade simulation task evaluates a model’s ability to synthesize physically correct and temporally consistent urban shade patterns under specified solar and temporal conditions. For each spatial tile, the model takes as input (i) an aligned image or its corresponding building skeleton representation, together with (ii) a text prompt $T$ encoding solar geometry attributes, such as examples in Section~\ref{sec:data}. The objective is to generate a shaded image $\hat{X}_{i,t}$ that matches the ground-truth shade dynamics $X_{i,t}$ produced by physically based simulation. 

\vspace{1mm}
\noindent\textbf{Baselines.}
The methods we compare include a vanilla Diffusion model~\cite{rombach2022high}, ControlNet~\cite{zhang2023adding}, and an Edge-conditioned ControlNet variant and DeepShade~\cite{da2025deepshade}.
All models are trained for 50 epochs, and each experiment is repeated five times to report the mean and standard deviation.

\vspace{1mm}
\noindent\textbf{Evaluation Protocol.}
We evaluate generative shade simulation quality using complementary metrics:  Structural Similarity Index Measure (SSIM~$\uparrow$)~\cite{mudeng2022prospects} and Learned Perceptual Image Patch Similarity (LPIPS~$\downarrow$)~\cite{ghazanfari2023r}. These metrics jointly assess pixel accuracy, structural consistency, region overlap, boundary alignment, and perceptual similarity between the generated shade image $\hat{X}_{i,t}$ and the ground truth $X_{i,t}$.

\vspace{1mm}
\noindent\textbf{Results.}
Quantitative results are reported in Figure~\ref{fig:result1}, using SSIM and LPIPS as evaluation metrics. Across both dense and sparse urban scenarios, the DeepShade~\cite{da2025deepshade} method consistently outperforms all baselines, achieving higher structural similarity and lower perceptual distance on both in-domain and out-of-domain datasets. These results indicate that the proposed generative framework captures not only fine-grained shade boundaries but also the underlying temporal and geometric regularities governing shade formation, leading to robust generalization across diverse cities and environmental conditions.

\subsection{Shade Segmentation}

% \noindent\textbf{Task Objective.}
The shade segmentation task evaluates the ability of segmentation models to accurately delineate shaded regions under varying solar conditions and visual contexts.
Unlike conventional object segmentation, urban shade exhibits fine-grained, low-contrast, and temporally varying boundaries that are strongly influenced by solar geometry and urban layout.
This task assesses whether existing segmentation models can robustly identify such nuanced shade patterns, which is a critical prerequisite for downstream tasks such as heat-aware routing and exposure analysis.

\begin{table*}[t!]
\centering
\caption{ The performance table for the task of segmenting shades by Simulated Image (Left) vs. Satellite Image (Right). Performance comparison of different models across multiple metrics at different times of the day.}
\label{tab:model_performance_combined}
    \vspace{-2mm}
\resizebox{\textwidth}{!}{
\begin{tabular}{c||c}
    % Left Table: Simulated Image
    \begin{tabular}{llcc|cc|cc}
    \toprule
    \multicolumn{8}{c}{\textbf{Simulated Image}} \\
    \midrule
    \multirow{2}{*}{\textbf{City Name}} & \multirow{2}{*}{\textbf{Model}} & \multicolumn{2}{c|}{\textbf{6 - AM}} & \multicolumn{2}{c|}{\textbf{12 - PM}} & \multicolumn{2}{c}{\textbf{6 - PM}} \\
    \cline{3-8}
     & & \textbf{DICE $\uparrow$} & \textbf{IoU $\uparrow$} & \textbf{DICE $\uparrow$} & \textbf{IoU $\uparrow$} & \textbf{DICE $\uparrow$} & \textbf{IoU $\uparrow$} \\
    \midrule
    \multirow{3}{*}{Phoenix} 
        & SAM & 0.9824 & 0.9701 & 0.9981 & 0.9962 & 0.9869 & 0.9903 \\
        & SAM2 & 0.9919 & 0.9844 & 0.9971 & 0.9943 & 0.9933 & 0.9905 \\
        & GeoSAM & 0.9741 & 0.9731 & 0.9044 & 0.9575 & 0.9905 & 0.9904 \\
    \midrule
    \multirow{3}{*}{Mexico City} 
        & SAM & 0.979 & 0.964 & 0.996 & 0.9921 & 0.9811 & 0.9812 \\
        & SAM2 & 0.9854 & 0.9719 & 0.9917 & 0.9838 & 0.9838 & 0.9865 \\
        & GeoSAM & 0.9745 & 0.9546 & 0.9625 & 0.9884 & 0.9884 & 0.9882 \\
    \midrule
    \multirow{3}{*}{Madrid} 
        & SAM & 0.9465 & 0.9367 & 0.9951 & 0.9902 & 0.8967 & 0.8948 \\
        & SAM2 & 0.9888 & 0.9785 & 0.9941 & 0.9883 & 0.9927 & 0.9925 \\
        & GeoSAM & 0.9567 & 0.9399 & 0.9815 & 0.9847 & 0.9883 & 0.9881 \\
    \midrule
    \multirow{3}{*}{Beijing} 
        & SAM & 0.969 & 0.9454 & 0.9962 & 0.9924 & 0.9868 & 0.9869 \\
        & SAM2 & 0.9901 & 0.9807 & 0.9961 & 0.9923 & 0.9729 & 0.9728 \\
        & GeoSAM & 0.9714 & 0.9535 & 0.9869 & 0.9901 & 0.9657 & 0.9658 \\
    \bottomrule
    \end{tabular}
    &
    % Right Table: Satellite Image
    \begin{tabular}{llcc|cc|cc}
    \toprule
    \multicolumn{8}{c}{\textbf{Satellite Image}} \\
    \midrule
    \multirow{2}{*}{\textbf{City Name}} & \multirow{2}{*}{\textbf{Model}} & \multicolumn{2}{c|}{\textbf{6 - AM}} & \multicolumn{2}{c|}{\textbf{12 - PM}} & \multicolumn{2}{c}{\textbf{6 - PM}} \\
    \cline{3-8}
     & & \textbf{DICE $\uparrow$} & \textbf{IoU $\uparrow$} & \textbf{DICE $\uparrow$} & \textbf{IoU $\uparrow$} & \textbf{DICE $\uparrow$} & \textbf{IoU $\uparrow$} \\
    \midrule
    \multirow{3}{*}{Phoenix} 
        & SAM & 0.2732 & 0.1895 & 0.4734 & 0.348 & 0.2783 & 0.1908 \\
        & SAM2 & 0.3711 & 0.2615 & 0.3383 & 0.246 & 0.2569 & 0.1752 \\
        & GeoSAM & 0.294 & 0.1901 & 0.4241 & 0.3701 & 0.2954 & 0.1911 \\
    \midrule
    \multirow{3}{*}{Mexico City} 
        & SAM & 0.3394 & 0.2477 & 0.4674 & 0.6561 & 0.3577 & 0.6236 \\
        & SAM2 & 0.3522 & 0.2401 & 0.2715 & 0.1756 & 0.4113 & 0.2809 \\
        & GeoSAM & 0.3219 & 0.23129 & 0.1414 & 0.6085 & 0.3831 & 0.4503 \\
    \midrule
    \multirow{3}{*}{Madrid} 
        & SAM & 0.0245 & 0.0133 & 0.1009 & 0.0587 & 0.0658 & 0.0376 \\
        & SAM2 & 0.1274 & 0.0814 & 0.0554 & 0.031 & 0.1892 & 0.1242 \\
        & GeoSAM & 0.0983 & 0.1349 & 0.081 & 0.0319 & 0.1493 & 0.104 \\
    \midrule
    \multirow{3}{*}{Beijing} 
        & SAM & 0.4734 & 0.3453 & 0.6041 & 0.4474 & 0.5068 & 0.3741 \\
        & SAM2 & 0.5748 & 0.4225 & 0.5512 & 0.4 & 0.5134 & 0.3747 \\
        & GeoSAM & 0.4911 & 0.3664 & 0.4391 & 0.4013 & 0.4991 & 0.431 \\
    \bottomrule
    \end{tabular}
\end{tabular}
}
\end{table*}

\vspace{1mm}
\noindent\textbf{Baselines.}
We benchmark three representative segmentation models: the Segment Anything Model (SAM)~\cite{kirillov2023segment}, Segment Anything Model v2 (SAM2)~\cite{ravi2024sam2}, and GeoSAM~\cite{sultan2023geosam}, a variant fine-tuned on geospatial and infrastructure datasets.
For fair comparison, all models are evaluated in a zero-shot inference setting without additional fine-tuning, reflecting their out-of-the-box segmentation capability.

\vspace{1mm}
\noindent\textbf{Evaluation Protocol.}
To examine model robustness across visual domains and temporal conditions, we design two segmentation settings evaluated at three representative timestamps (6:00 AM, 12:00 PM, and 6:00 PM):
(i) segmenting shade regions from simulated shaded images, and
(ii) segmenting shade regions directly from real-world satellite images.
This dual-setting design allows us to analyze segmentation performance under idealized (simulation) and realistic (satellite) visual conditions, and to quantify the domain gap between them.
Segmentation quality is evaluated using the Dice coefficient ($\uparrow$) and Intersection over Union (IoU, $\uparrow$), following standard practice in segmentation benchmarks~\cite{bertels2019optimizing, van2019deep, zhao2024foundation}.
Metric definitions and implementation details are provided in the appendix.

\begin{figure}[h!]
    \centering
    \includegraphics[width=0.48\textwidth]{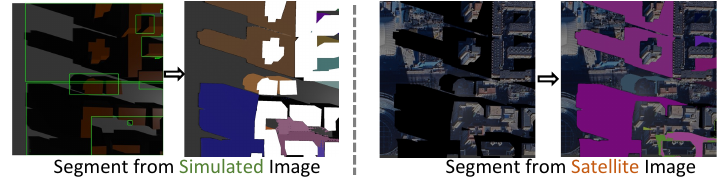}
    \vspace{-5mm}
    \caption{ The two types of segmentation as designed in the experiment. The segmentation from the simulation snapshot image is easier, and the segmentation from the satellite image is much harder.}
    \label{fig:seg}
    \vspace{-2mm}
\end{figure}

\vspace{1mm}
\noindent\textbf{Results.}
Experimental results in Table~\ref{tab:model_performance_combined} show that while foundation segmentation models can capture coarse shaded regions under favorable conditions, their performance degrades noticeably for fine-grained and low-contrast shade boundaries, particularly in satellite imagery and during early morning or late afternoon when shadows are elongated.
Overall, these findings highlight the challenges of urban shade segmentation and underscore the need for physically grounded datasets such as \oursdata to support more robust and temporally consistent shade perception.
\subsection{3D Reconstruction}

% \noindent\textbf{Task Objective.}
The 3D reconstruction task evaluates a model’s ability to recover building-level geometric structure from 2D observations.
Given an aligned satellite image $X_j'$ (and optionally its building topology representation $X_i$), the goal is to reconstruct a 3D building geometry that is consistent with the true urban structure.
Unlike purely visual benchmarks, \oursdata provides physically grounded 3D supervision via Blender-exported meshes, enabling geometry-aware evaluation rather than qualitative inspection alone.
For each spatial tile $i$, the reconstruction model takes as input an aligned satellite image $X_j'$ and/or its corresponding building skeleton image $X_i$.
The output is a reconstructed 3D shape $\hat{M}_i$, represented as either a mesh or a point cloud.
Ground-truth geometry is provided by the Blender-exported building mesh $M_i$, which is derived from OSM topology and physically consistent simulation.
This explicit pairing $(X_j', X_i, M_i)$ allows direct supervision and quantitative evaluation of reconstruction quality.

\begin{figure}[h!]
    \centering
    \includegraphics[width=0.48\textwidth]{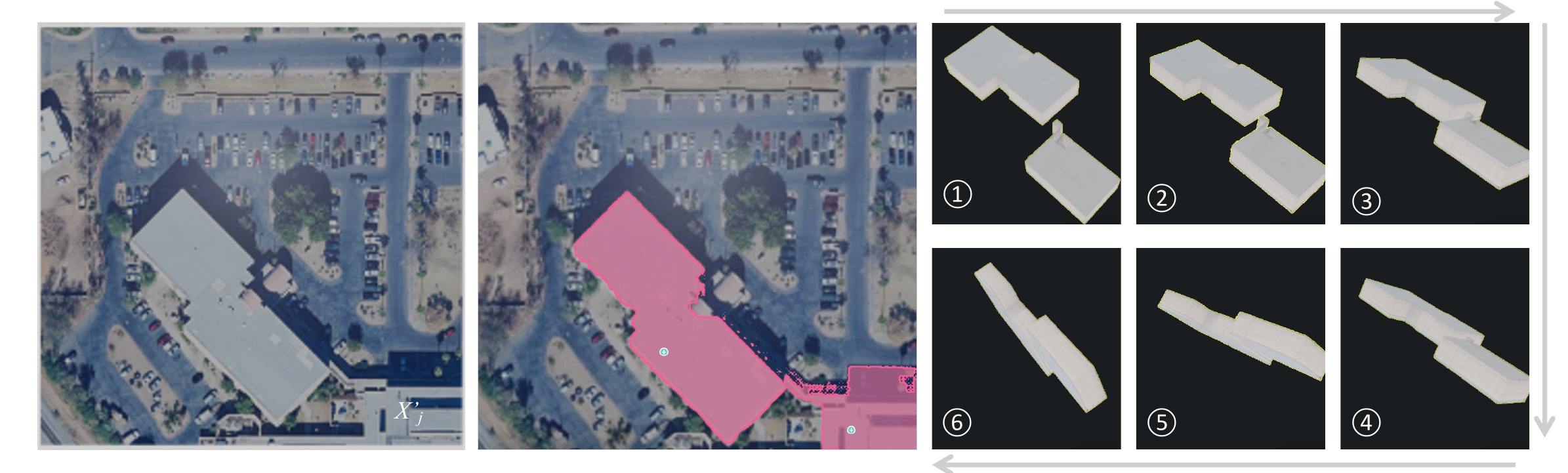}
    \vspace{-6mm}
    \caption{ Illustration of the 3D building reconstruction task enabled by \oursdata. 
From left to right: an aligned satellite image after dataset alignment, the corresponding building prompt/topology used to condition reconstruction, and multiple reconstructed 3D building meshes rendered from different viewpoints. 
The example demonstrates how spatially aligned imagery and building topology provided by \oursdata can support geometry-aware reconstruction under realistic urban conditions.}
    \label{fig:3d}
\end{figure}

\vspace{1mm}
\noindent\textbf{Baselines.}
As a proof of concept in Figure~\ref{fig:3d}, we demonstrate this task using a representative off-the-shelf 3D reconstruction model, \textit{SAM-3D}, applied to aligned inputs from \oursdata.
We emphasize that this experiment serves as a diagnostic demonstration rather than an exhaustive benchmark.
The primary contribution of \oursdata lies in enabling standardized evaluation for 3D urban reconstruction under realistic alignment and solar conditions, rather than proposing a new reconstruction model.

\vspace{1mm}
\noindent\textbf{Evaluation Protocol.}
To quantitatively assess reconstruction quality, one could adopt the \emph{Chamfer Distance} (CD), a widely used metric for comparing reconstructed and ground-truth geometry.
Given a reconstructed point set $\hat{P}_i$ sampled from $\hat{M}_i$ and a ground-truth point set $P_i$ sampled from $M_i$, the Chamfer Distance is defined as:
$
\mathrm{CD}(\hat{P}_i, P_i) =
\frac{1}{|\hat{P}_i|} \sum_{p \in \hat{P}_i} \min_{q \in P_i} \|p - q\|_2^2
+
\frac{1}{|P_i|} \sum_{q \in P_i} \min_{p \in \hat{P}_i} \|q - p\|_2^2 
$
Lower values indicate better geometric consistency between the reconstructed shape and the ground-truth building geometry.
Following prior work~\cite{wusupplementary}, this metric can be extended to density-aware variants to account for non-uniform point sampling on complex urban structures.

\vspace{1mm}
\noindent\textbf{Discussion.}
By providing aligned satellite imagery $X_j'$, building topology $X_i$, and high-quality ground-truth meshes $M_i$, \oursdata establishes a unified benchmark for urban-scale 3D reconstruction.
Future work can systematically compare different reconstruction paradigms: single-view, multi-view, or topology-guided under consistent geometric supervision.
This makes \oursdata a valuable testbed for studying how alignment quality, shading cues, and solar-driven appearance variations influence 3D urban reconstruction.

\section{Conclusion and Discussion}
This paper introduces \oursdata, a large-scale benchmark for urban shade understanding that integrates aligned satellite imagery, physically grounded solar modeling, and multi-task evaluation protocols. The benchmark supports three complementary tasks, including generative shade simulation, shade segmentation, and 3D building reconstruction, which enables systematic and reproducible evaluation under realistic urban and environmental conditions.
Our empirical results highlight both progress and remaining challenges. While generative models can learn meaningful correlations between urban geometry, solar conditions, and shade evolution, accurately simulating fine-grained and temporally consistent shade patterns remains difficult, especially under complex building layouts and extreme illumination conditions.
Similarly, shade segmentation proves to be a non-trivial perceptual task: even strong foundation models struggle to robustly delineate soft shadow boundaries and subtle occlusion patterns across different times of day and seasonal settings.
These findings suggest that shade is a fundamentally challenging visual–physical phenomenon that requires tighter integration of geometry, physics, and perception.

\vspace{1mm}
\noindent \textbf{Limitations and future work:} Although \oursdata handles missing buildings through alignment, the generative refinement strategy may replace removed buildings with plausible surfaces, such as parking lots or open ground, that are not always identical to the true real-world appearance
Additionally, the current dataset models shade variation on an hourly, discrete time grid, which captures broad temporal dynamics but does not yet fully represent continuous or finer-grained shade changes.  
\oursdata also opens several research directions: such as studying hybrid models that more tightly couple physical constraints with learning-based representations, improving robustness to appearance variations, and jointly reasoning about geometry, illumination, and semantics.

\section*{GenAI Disclosure}
Generative AI tools were used in a limited and auxiliary manner, specifically for typographical review and grammar correction and for part of image generation as a technical step required in Eq.(~\ref{eq:genImage}). All core components of this work, including the simulation framework, dataset construction, alignment and preprocessing pipeline, benchmark design, experimental implementation, and analysis, were conceived, designed, and implemented by the researchers. 

% %%
% %% The acknowledgments section is defined using the "acks" environment
% %% (and NOT an unnumbered section). This ensures the proper
% %% identification of the section in the article metadata, and the
% %% consistent spelling of the heading.
\begin{acks}
The work was partially supported by NSF awards \#2442477 and \#2550203. We thank the Amazon Research Awards, Cisco Faculty Research Awards, and Toyota Faculty Research Awards. The authors acknowledge the Google Cloud Research Credits Program for providing us with API credits and the Research Computing at Arizona State University for providing computing resources. 
The views and conclusions in this paper are those of the authors and should not be interpreted as representing any funding agencies.
\end{acks}

\bibliographystyle{ACM-Reference-Format}
\bibliography{sample-base}

\clearpage
\appendix

\section{Dataset Details}
\label{sec:datasetdetail}

\subsection{Cities covered in \oursdata}
\label{sec:dataset1}

\begin{figure}[h]
    \centering
    \includegraphics[width=0.5\textwidth]{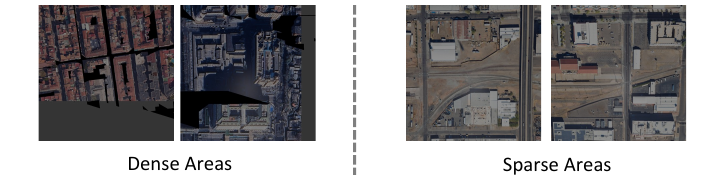}
    \caption{The example images from dense (Beijing and Madrid) and sparse areas (Tempe).}
    \label{fig:sparseDense}
\end{figure}

As shown in Figure~\ref{fig:dims} and Table~\ref{tab:dataset-coverage}, \oursdata spans six continents, covering diverse urban layouts (sparse vs. dense), and traffic rules (left-hand vs. right-hand driving), and includes at least three distinct countries per category to ensure broad geographic and infrastructural representation. An example comparison of these urban layouts is shown in Figure~\ref{fig:sparseDense}.

% \begin{figure}[h!]
%     \centering
%     \includegraphics[width=0.45\textwidth]{figs/Datadims.pdf}
%     \vspace{-4mm}
%     \caption{The multiple dimensions of the \oursdata.}
%     \label{fig:dims}
%     \vspace{-5mm}
% \end{figure}

\begin{table}[h!]
\centering
\caption{Cities included in \oursdata}
\label{tab:dataset-coverage}
\resizebox{0.48\textwidth}{!}{ % Adjust table size
\begin{tabular}{llcc}
\toprule
\textbf{Continent} & \textbf{Country} & \textbf{Dense Building City} & \textbf{Sparse Building City} \\ 
\midrule
\multirow{3}{*}{North America} 
    & USA & Phoenix & Tempe \\
    & Canada & Toronto & Calgary \\
    & Mexico & Mexico City & Tulum \\
\midrule
\multirow{3}{*}{South America} 
    & Brazil & São Paulo & Brasília \\
    & Argentina & Buenos Aires & Salta \\
    & Chile & Santiago & Calama \\
\midrule
\multirow{3}{*}{Europe} 
    & Spain & Madrid & Seville \\
    & France & Paris & Nîmes \\
    & Italy & Rome & Sardinia \\
\midrule
\multirow{3}{*}{Africa} 
    & Nigeria & Lagos & Abuja \\
    & Egypt & Cairo & Aswan \\
    & South Africa & Johannesburg & Cape Town \\
\midrule
\multirow{3}{*}{Asia} 
    & China & Beijing & Xi'an \\
    & Japan & Tokyo & Nagoya \\
    & India & Mumbai & Jaipur \\
\midrule
\multirow{2}{*}{Oceania} 
    & Australia & Sydney & Outback \\
    & New Zealand & Auckland & Chatham \\
\bottomrule
\end{tabular}
} % End of resizebox
\end{table}

\subsection{Implementation Details}

\subsubsection{Rationale of Traffic Rule Variation}
The inclusion of traffic rule variation (left-hand vs. right-hand driving) as a dataset dimension is essential due to its influence on urban layouts and shade distributions, and its potential for further research topics.
Traffic rules shape the design of roads, intersections, and public infrastructure, which directly affects the positioning and orientation of buildings, sidewalks, and greenbelts. For example, left-hand driving countries often place bus stops and pedestrian pathways on the left, creating distinct shade patterns compared to right-hand driving countries. Additionally, the allocation of greenery and roadside infrastructure differs, impacting both temporary and permanent shading. By accounting for these variations, the dataset ensures that models can capture the nuanced relationships between traffic systems and shade distributions, improving their generalization across diverse urban environments.

\subsubsection{Resolution Calculation}\label{sec:reso}
At Google Maps tile level 13, the resolution at the equator is approximately 19.11 meters per pixel. This resolution is determined based on the Earth's circumference and the hierarchical tiling system used by Google Maps, where each zoom level increases the number of tiles exponentially. At this level, a single pixel represents a ground distance of around 19.11 meters, providing a medium-scale view suitable for regional mapping and analysis.

\subsubsection{Post Processing}
\begin{figure}[h!]
    \centering
    \includegraphics[width=0.99\linewidth]{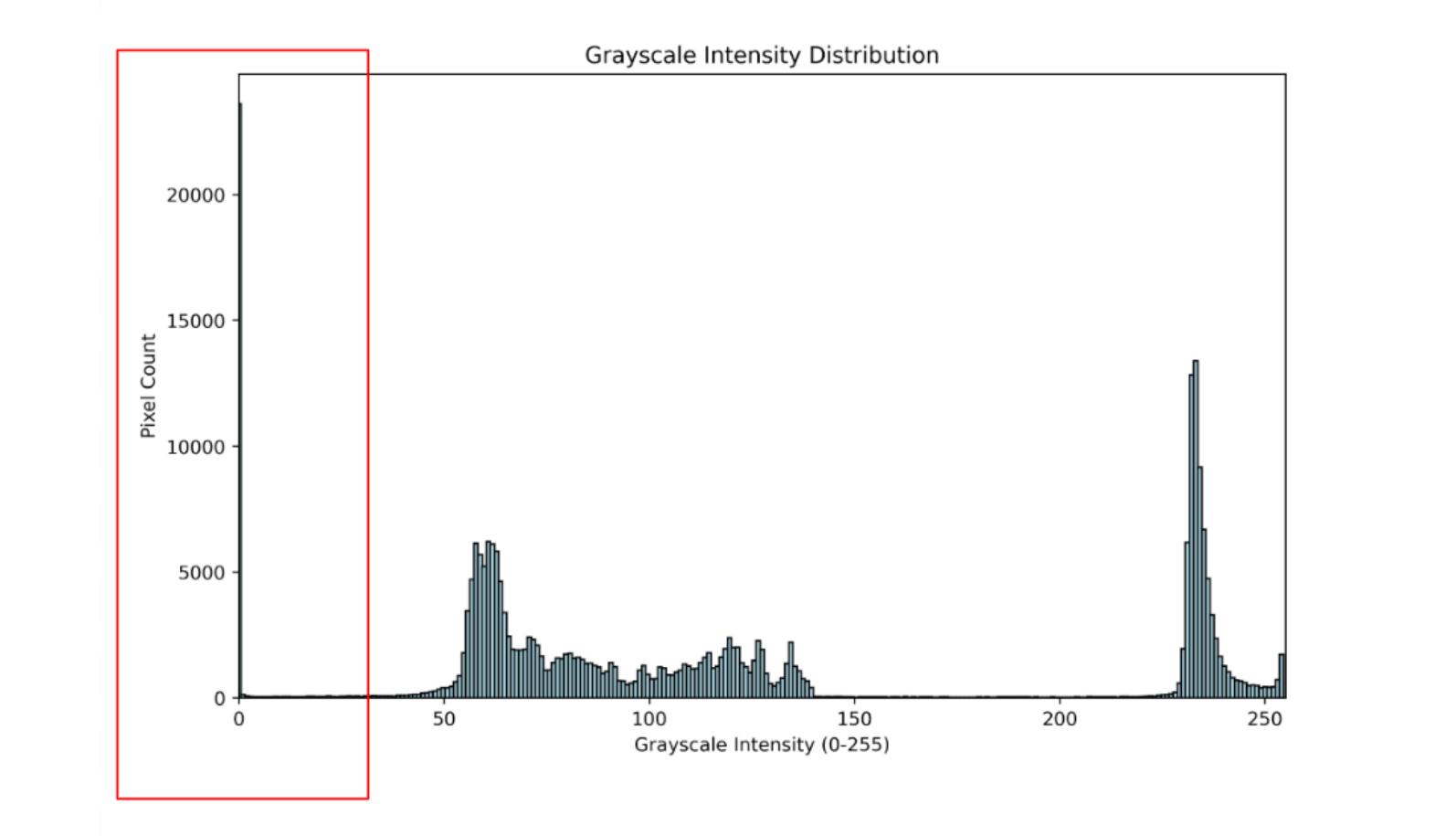}
    \caption{The empirical value for setting the shade threshold when conducting the filtering.}
    \label{fig:threshold}
\end{figure}

To benefit the deep learning models, we need a ground-truth shade for measuring the model's output, and provide feedback through loss functions. To obtain the actual ground truth shade, we take another snapshot of the scene without sunshine, named as `skeleton image' $x^{sk}$, which only provides a pure building structure, then the pure shade can be obtained through the formula:
\begin{equation}\label{eq:threhsold}
    x^{gt} = x^{shade} - x^{sk} - \mathbb{I}(x^{shade} \leq \alpha)
\end{equation}
where \( x^{gt} \) represents the extracted ground truth shade, \( x^{shade} \) is the shaded snapshot taken under sunlight, capturing the shades, sides, and full building observation, and \( x^{sk} \) is the skeleton snapshot taken without sunlight, providing only the pure building structure. The term \( \mathbb{I}(x^{shade} \leq \alpha) \) is an indicator function that removes low-intensity grey values (noise thresholding), where \( \alpha \) is a predefined threshold (chosen as 30 in this paper) as shown in Figure~\ref{fig:threshold}. This equation effectively removes structural elements and noise, isolating the actual shaded areas to provide precise ground truth data for machine learning models.

\subsection{The Metrics Details}
\subsubsection{The Metrics Used for Generative Shade Simulation}
\paragraph{Mean Squared Error (MSE) \(\downarrow\)}
MSE quantifies the pixel-wise difference between the generated and ground truth images, defined as: $\text{MSE} = \frac{1}{N} \sum_{i=1}^{N} (x_i - y_i)^2$,
where \(x_i\) and \(y_i\) are the pixel values of the generated and ground truth images, respectively, and \(N\) is the total number of pixels. The range of MSE is \([0, \infty)\), where lower values indicate better similarity.

\paragraph{Mean Intersection over Union (mIoU) \(\uparrow\)}
mIoU evaluates the overlap between the binary shadow masks of the generated and ground truth images. It is calculated as:
\begin{equation}
    \text{mIoU} = \frac{\text{Intersection}}{\text{Union}} = \frac{|A \cap B|}{|A \cup B|}
\end{equation}

where \(A\) and \(B\) represent the binary shadow masks. The range of mIoU is \([0, 1]\), where higher values indicate better overlap.

\paragraph{Structural Similarity Index Measure (SSIM) \(\uparrow\)}
SSIM measures the perceptual similarity between the generated and ground truth images~\cite{wess1994using}. The range of SSIM is \([0, 1]\), where a higher value indicates better structural similarity.

%, considering luminance, contrast, and structure. It is computed as:
% \[
% \text{SSIM}(x, y) = \frac{(2\mu_x \mu_y + C_1)(2\sigma_{xy} + C_2)}{(\mu_x^2 + \mu_y^2 + C_1)(\sigma_x^2 + \sigma_y^2 + C_2)},
% \]
% where \(\mu_x\), \(\mu_y\) are the mean intensities, \(\sigma_x^2\), \(\sigma_y^2\) are variances, \(\sigma_{xy}\) is the covariance, and \(C_1, C_2\) are constants to stabilize division. 

\subsubsection{The Metrics for Shade Segmentation Tasks}

\paragraph{Dice Similarity Coefficient (DICE) \(\uparrow\)}
DICE measures the overlap between the predicted and ground truth segmentations, defined as:
\begin{equation}
    \text{DICE} = \frac{2 \cdot |X \cap Y|}{|X| + |Y|}
\end{equation}

where \(X\) and \(Y\) represent the sets of pixels belonging to the predicted and ground truth regions, respectively. The range of DICE is \([0,1]\), where higher values indicate better similarity, with 1 representing perfect overlap.

\paragraph{Intersection over Union (IoU) \(\uparrow\)}
IoU measures the similarity between the predicted and ground truth segmentation regions, defined as:
\begin{equation}
    \text{IoU} = \frac{|X \cap Y|}{|X \cup Y|}
\end{equation}

where \(X\) and \(Y\) are the predicted and ground truth segmentation regions, respectively. The range of IoU is \([0,1]\), where higher values indicate better similarity, with 1 representing a perfect match.

\begin{figure*}[h]
    \centering
    \includegraphics[width=0.8\textwidth]{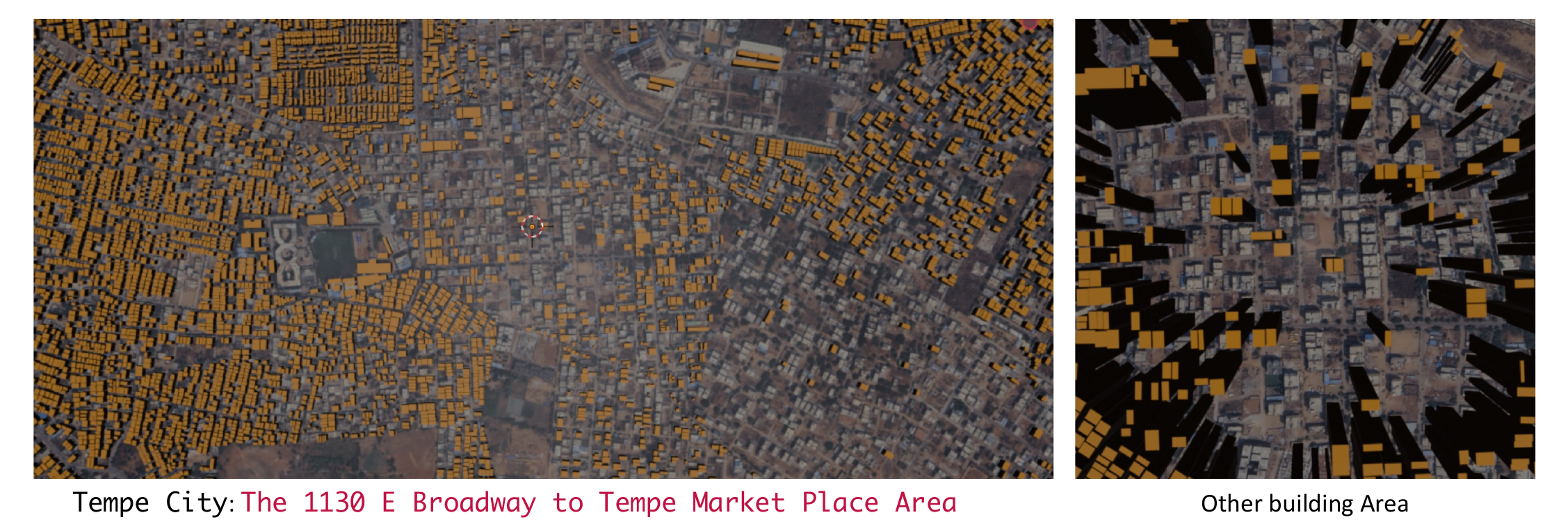}
    \caption{An example case of missing buildings from the OSM map.}
    \label{fig:missing}
\end{figure*}

\begin{figure*}[h!]
    \centering
    \includegraphics[width=0.99\linewidth]{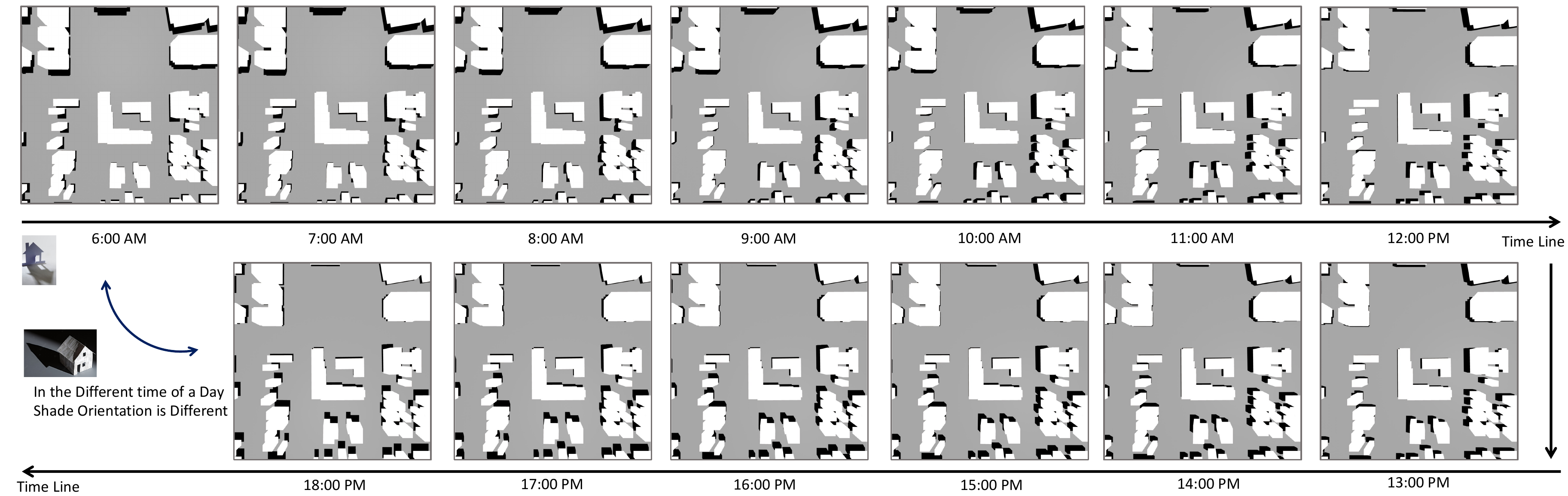}
    \caption{Example from \oursdata dataset illustrating temporal shade evolution in a Northern Hemisphere urban area (Tempe, AZ, USA). The sequence shows top-down shade maps at hourly intervals, highlighting how shadow orientation and coverage change throughout the day as the solar position varies.}
    \label{fig:north}
\end{figure*}

\begin{figure*}[h!]
    \centering
    \includegraphics[width=0.99\linewidth]{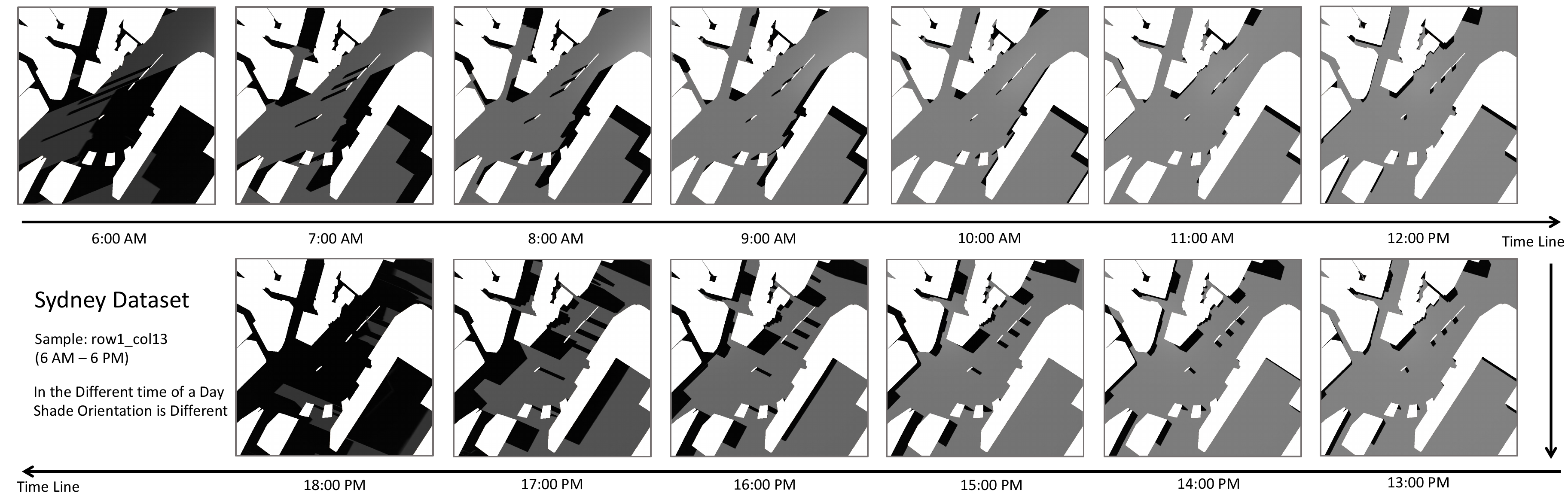}
    \caption{Example from \oursdata dataset illustrating temporal shade evolution in a Southern Hemisphere urban area (Sydney, Australia). The hourly top-down shade maps show how shadow orientation and coverage change throughout the day, exhibiting patterns opposite to those observed in the Northern Hemisphere due to differing solar geometry.}
    \label{fig:south}
\end{figure*}

\end{document}